\documentclass[conference]{IEEEtran}
\usepackage{times}

\usepackage[numbers]{natbib}
\usepackage{multicol}
\usepackage[bookmarks=true]{hyperref}
\usepackage{xspace}
\usepackage{multirow}
\usepackage{amssymb}
\usepackage{graphicx}
\usepackage{subfigure}
\usepackage{xcolor}
\usepackage{makecell}
\usepackage{stmaryrd}
\usepackage{pifont}

\usepackage{amsmath}

\begin{document}

\title{\sysname: Transferring Visuomotor Policies From Static Data Training to Dynamic Robot Manipulation}




%
\author{\authorblockN{Yifan Duan\authorrefmark{1}$^1$,
Heng Li\authorrefmark{1}$^1$,
Yilong Wu\authorrefmark{1}$^1$, 
Wenhao Yu$^2$,
Xinran Zhang$^1$,
Yedong Shen$^1$,\\
Jianmin Ji$^1$ and Yanyong Zhang\authorrefmark{2}$^3$}
\authorblockA{$^1$School of Computer Science and Technology, University of
Science and Technology of China, Hefei, China}
\authorblockA{$^2$Institute of Advanced Technology, University of
Science and Technology of China, Hefei, China}
\authorblockA{$^3$School of Artificial Intelligence and Data Science, University of
Science and Technology of China, Hefei, China}

\authorblockA{\authorrefmark{1}Equal Contribution~\authorrefmark{2}Corresponding Author}
\authorblockA{Project page: \url{https://yjsx.top/stdarm}}

}

\newcommand{\dnote}[1]{ \color{purple}[Yifan: #1]  \color{black}}
\newcommand{\etal}{\textit{et al.}\@}
\newcommand{\sysname}{\textit{STDArm}\xspace}
\newcommand{\sysnameA}{\textit{MT-3DoF}\xspace}
\newcommand{\sysnameB}{\textit{MF-5DoF}\xspace}
\newcommand{\sysnameC}{\textit{UAV-3DoF}\xspace}
\newcommand{\sysnameD}{\textit{LS-3DoF}\xspace}
\maketitle

\begin{abstract}

Recent advances in mobile robotic platforms like quadruped robots and drones have spurred a demand for deploying visuomotor policies in increasingly dynamic environments. However, the collection of high-quality training data, the impact of platform motion and processing delays, and limited onboard computing resources pose significant barriers to existing solutions. In this work, we present \sysname, a system that directly transfers policies trained under static conditions to dynamic platforms without extensive modifications.

The core of \sysname is a real-time action correction framework consisting of: (1) an action manager to boost control frequency and maintain temporal consistency, (2) a stabilizer with a lightweight prediction network to compensate for motion disturbances, and (3) an online latency estimation module for calibrating system parameters. In this way, \sysname achieves centimeter-level precision in mobile manipulation tasks.

We conduct comprehensive evaluations of the proposed \sysname on two types of robotic arms, four types of mobile platforms, and three tasks. Experimental results indicate that the \sysname enables real-time compensation for platform motion disturbances while preserving the original policy's manipulation capabilities, achieving centimeter-level operational precision during robot motion.
\end{abstract}

\IEEEpeerreviewmaketitle

\section{Introduction}
Visuomotor policies, which directly map visual observations to robot actions, have emerged as an effective paradigm for complex task acquisition~\cite{urain2024deep}. By integrating end-to-end imitation learning frameworks trained on human demonstrations, visuomotor policies eliminate the need for task-specific modular pipeline design, translating raw sensor input into robot actions without intermediate representations~\cite{rahmatizadeh2018vision}. Recent advances have enabled robotic systems with remarkable capabilities such as battery insertion~\cite{ACT}, food preparation~\cite{dp}, and textile handling~\cite{weng2022fabricflownet}, etc.

The acquisition of high-quality human demonstrations, typically achieved through teleoperation, plays a pivotal role in the successful skill acquisition using imitation learning-based visuomotor policies~\cite{seo2023deep}. This process is relatively simple with stationary robotic systems, leading to common experimental setups that use table-mounted manipulators. 
However, significant challenges arise when working with dynamic robot platforms, such as  drones, humanoid or quadruped~\cite{umi_on_legs,humanplus}. 
These robots exhibit both active movement for navigation and wobble caused by their instability.
The inherent mobility of these platforms requires skilled operators who can maintain stable action outputs while compensating for base movement disturbances during data collection. 
This transforms what would otherwise be a straightforward data collection step into potentially the most challenging part of the entire process.

\begin{figure}[t]
	\centering
	\includegraphics[width = .9\linewidth]{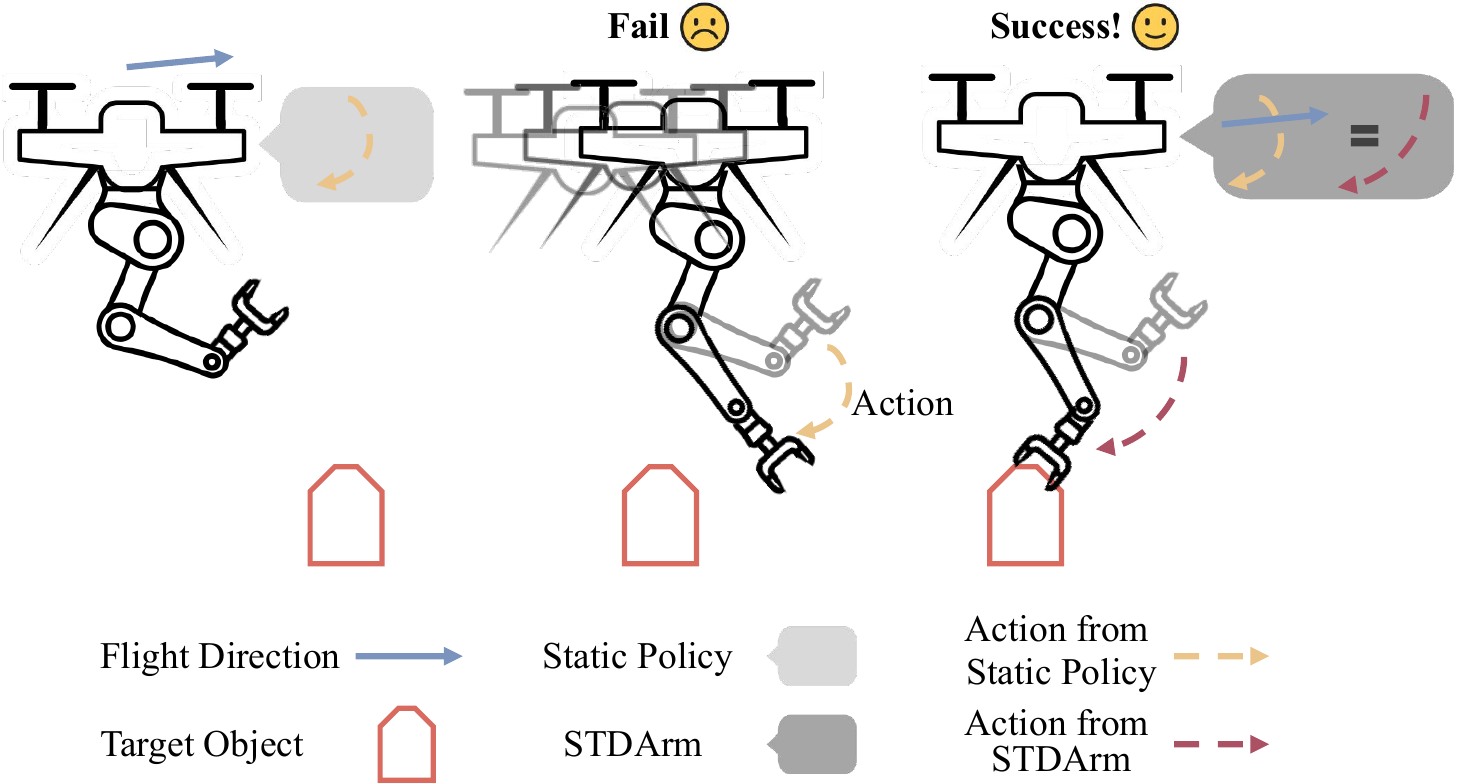}
	\caption{When deploying a policy trained on static data on computationally constrained edge devices, action drift may occur due to delays like inference and execution latency. Our system, \sysname, addresses this by direct rectification of actions, enabling reliable task completion without policy modifications.}
	\label{fig:first}
\end{figure}
Beyond the challenges in data collection, directly deploying policies trained in stationary conditions on dynamic platforms is ineffective. First, performing policy inference on low-computational platforms carried by small robots often results in significant processing delays, causing the generated actions to become outdated due to the robot's motion, as shown in Fig.~\ref{fig:first}. Second, current visuomotor policies~\cite{ACT,dp,3ddp} commonly employ action chunking to promote temporal action consistency, which involves generating multiple actions in a single inference for sequential execution. 
This approach further reduces the robot's responsiveness to environmental changes. Lastly, even if above challenges are overcome, policies trained on specific mobile robot struggle to generalize across different robotic systems due to their  distinct locomotion characteristics.

To address the aforementioned challenges, we develop \sysname, which transfers visuomotor policy from \underline{\textbf{s}}tatic data training \underline{\textbf{t}}o \underline{\textbf{d}}ynamic platforms with negligible cost, equipping various mobile robots with robust capabilities inherited from imitation learning.
Our key insight is that, in addition to visual and arm pose inputs, \sysname incorporates additional high-frequency observations of the robot's position and employs a lightweight network to perform real-time action corrections based on the robot's movement. 
By increasing the control frequency, the system achieves swift responsiveness to changes in robot's state.
Specifically, an action manager module is first designed to decouple the policy network's asynchronous inference from the real-time control loop, while interpolating the policy outputs to higher frequency. This establishes high-frequency control as the foundation of the entire system.
Building on this, \sysname combines high-frequency ego-motion tracking through a visual SLAM system – operating at 10× the base policy's observation rate – with a lightweight neural network that bridges the perception-action latency gap. This integrated solution enables real-time compensation for platform motion disturbances while maintaining the original policy's manipulation capabilities, achieving centimeter-level operational precision during mobile manipulation tasks.
Finally, to align \sysname's optimization objectives with task execution goals, we implement an action-stabilizing warm-up routine prior to mission execution, enabling online calibration of system parameters through controlled motion sequences.

To validate \sysname's effectiveness, we conduct experiments across two distinct robotic arm configurations and four mobile platforms, including drones. Two canonical visuomotor policies, diffusion policy (DP)~\cite{dp} and ACT~\cite{ACT}, serve as the foundational policy networks. We design three manipulation tasks with progressively stringent precision requirements to evaluate \sysname's manipulation stability during robot's movement. The most challenging task demands end-effector positioning accuracy within 4 cm. Experimental results demonstrate that while the native policy achieves near-zero success rates during the robot's movement, \sysname attains success rates comparable to those of the native policy under ideal static conditions.
These experimental configurations validate that \sysname enables cost-effective migration of static-trained policy networks to diverse dynamic platforms.

Overall, our contributions are as follows:
\begin{itemize} 
    \item We present an efficient system, \sysname, enabling direct deployment of static-trained policy networks to mobile robotic platforms with negligible cost, while preserving the policy's original performance.  
    \item The designed action manager module is incorporated to enhance robotic decision-making and control frequencies, forming the foundational architecture of \sysname.
    \item By predicting the robot's state and estimating system latency online, we precisely maintain the stability of manipulation during the robot's movement.
    \item To validate our approach, we conduct experiments using two types of robotic arms across three different tasks.
    Furthermore, we deploy \sysname on a drone in a real-world scenario to validate its effectiveness.
\end{itemize}

\section{Related Work}
\subsection{Imitation Learning}
Imitation learning enables robots or agents to rapidly acquire complex behaviors by leveraging expert-provided demonstrations. In recent years, this approach has been widely applied to tasks such as robot manipulation~\cite{ACT,dp,OpenVLA,black2024pi_0,RDT,CARP,FAST}, navigation~\cite{LDP, nomad}, and mobile manipulation~\cite{mobile_aloha,umi_on_legs,humanplus}. 
Diffusion Policy~\cite{dp} leverages the robust multi-modal modeling capabilities of diffusion models to better address the challenges posed by the diversity in robot manipulation demonstrations. $\pi_0$~\cite{black2024pi_0} integrates a pre-trained multimodal language model with a flow-matching-based diffusion action head to achieve efficient and generalized robotic policies. CARP~\cite{CARP} aims to design an autoregressive generation policy that extracts multi-scale representations of the entire robot action sequence, generating action sequence predictions through a coarse-to-fine autoregressive process.
The efficient collection of high-quality expert data is one of the crucial prerequisites for the effectiveness of the aforementioned imitation learning methods. However, unstable and dynamic robotic platforms, like drones and legged robots, introduce significant challenges to the data collection process.
To address this, \sysname introduces an efficient system that allows policies trained on data collected from static platforms to be seamlessly transferred for inference on unstable and dynamic robotic platforms. 

The dynamically unstable platform also imposes requirements on the inference frequency of the policy, as a higher inference speed can better adapt to the dynamic changes in the task settings. In particular, the recently popular diffusion policy faces challenges in improving inference frequency due to its need for multiple iterative denoising steps. Consistency Policy~\cite{cp,manicm} attempts to use consistency distillation to enable a student network to learn a direct mapping over longer step spans on the probability flow ODE. AdaFlow~\cite{hu2024adaflow}, based on flow-based generative models, adaptively adjusts the computational cost during inference according to the variance of the state, thereby significantly improving inference efficiency while maintaining decision diversity. The above methods all attempt to reduce the number of denoising steps in diffusion policy to increase the overall policy inference frequency. 
Meanwhile, BID~\cite{bid} creates action chunks at each observation stamp and uses two loss functions to select chunks that try achieving both long-term consistency and short-term reactivity.
However, \sysname adopts a more direct approach by using an action manager and interpolated
action compensation, ensuring swift responses to state changes even on low-computational platforms.

Theoretically, the system is designed to be compatible with a wide range of learning-based algorithms, ensuring flexibility and broad applicability.


\begin{figure*}[t]
	\centering
	\includegraphics[width = .9\linewidth]{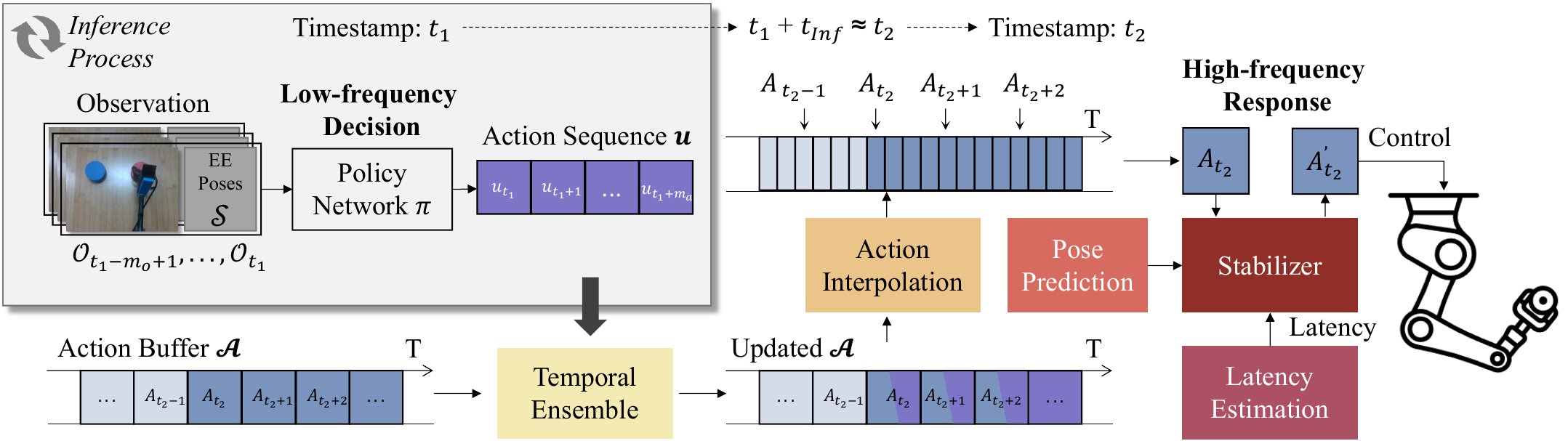}
	\caption{Pipeline of  \sysname. Given images and joint state observations,  our system first generates a low-frequency action sequence via a policy network. These actions are then passed to action manager, which maintains an action buffer using temporal ensemble and generates high-frequency actions through interpolation. Subsequently, a stabilizer refines the action based on estimated latency and real-time pose predictions to compensate for the motion of the robot platform. Finally, the stabilized action is sent to the robotic arm for precise and stable manipulation.}
	\label{fig:pipeline}
\end{figure*}

\subsection{Mobile Manipulation}
Early methods for mobile manipulation have shown notable success in tasks like grasping, including ground-based mobile manipulation~\cite{bajracharya2020mobile, chestnutt2005footstep, feng2014optimization, khatib1996force, wyrobek2008towards, rudin2022learning, fu2023deep} and unmanned aerial manipulation (UAM)~\cite{chen2022image, cao2023eso, wang2023millimeter}. However, these approaches often depend on human engineered designs tailored to specific scenarios or require extensive data collection for training, which restricts their adaptability across diverse engineering applications. To mitigate these limitations, MobileALOHA~\cite{mobile_aloha} introduces a cost-effective wheeled dual-arm system capable of remote operation to streamline data collection and relies on ACT~\cite{ACT} for policy learning and deployment. Despite its contributions, MobileALOHA is primarily designed for static operations within similar environmental conditions, limiting its broader applicability. Recognizing these constraints, UMI on Legs~\cite{umi_on_legs} focuses on efficient data collection and Diffusion Policy~\cite{dp} deployment across a variety of dynamic legged robots. Its performance is validated on platforms such as quadrupeds, demonstrating increased adaptability. 
However, when applied to highly dynamic platforms like drones, manipulation faces heightened challenges in achieving precise and robust control due to the dynamic and unstable nature of their bases. 
Addressing these challenges, our STDArm builds on the strengths of previous methods, enabling efficient data collection and stable deployment of imitation learning for manipulation operating on more challenging platforms, including drones. 
This advancement underscores the potential for enhanced control and adaptability in complex and dynamic environments.

\section{STDARM}

As shown in Fig.~\ref{fig:pipeline}, we present the \sysname architecture through its workflow, tracing the data flow from RGB image inputs and joint state measurements to final action execution.
We initially employ a foundational policy network to generate  action sequence for low-frequency decision-making, as detailed in Sec.~\ref{sec:system_overview:policy}.
Then, an action manager module enhances control frequency through temporal ensemble and action interpolation strategy, as discussed in Sec.~\ref{sec:system_overview:action_manager}.
To enable real-time compensation for platform motion disturbances, the stabilizer module, detailed in 
Sec.~\ref{sec:stabilizer}, facilitates the conversion of preliminary actions into the final executed actions.
Finally, we design an action-stabilizing warm-up routine before mission execution in Sec.~\ref{sec:latency_estimation}, to perform online parameter estimation essential for the stabilizer module.

It is worth mentioning that all these modules are deployed on edge devices without the assistance of external platforms for computation.


\subsection{Policy Network}
\label{sec:system_overview:policy}
 The policy network $\pi$ takes $m_o$ steps of observations $\mathcal{O}$, including images $\mathcal{I}$ and end-effector poses $\mathcal{S}$ as input. Based on these observations, the network predicts the action sequence $\mathbf{u}$ for the next $m_a$ time steps. Both the observation steps $m_o$ and prediction steps $m_a$ are predefined based on prior experience. Denoting the action sequence as $\mathbf{u}_{t+1:t+m_a}=\{u_{t+1},...,u_{t+m_a}\}$, this process can be formulated as:
\begin{equation}
\mathbf{u}_{t+1:t+m_a} = \pi(\mathcal{I}_{t-m_o+1:t}, \mathcal{S}_{t-m_o+1:t}),
\end{equation}
where $u$ is defined as the end-effector's position, which will be used to control the robotic arm. In order to avoid prolonged policy inference blocking the entire system, the policy network $\pi$ is assigned a dedicated subprocess. Following each inference cycle, the policy network immediately retrieves the latest observations to ensure a continuous generation of action sequences. This approach enhances the decision-making frequency of the policy network, which is beneficial for handling dynamic scenarios.

\subsection{Action Manager}
\label{sec:system_overview:action_manager}
To effectively select the appropriate actions for execution from the continuously generated action sequences, we design an action manager module to handle the output of the policy network. First, the action manager maintains an action buffer $\mathbf{\mathcal{A}}=\{A_t, t \in \mathbb{N}\}$, which sequentially stores both recently executed actions  and predicted actions awaiting execution. We define $A_t$ to be in the same form as $u$. A temporal ensemble algorithm is then employed to merge the newly generated action sequences $\mathbf{u}$ with the contents of the action buffer $\mathcal{A}$. Finally, action interpolation is applied to enhance the control frequency of the robotic arm.
The details are as follows:
\subsubsection{Temporal Ensemble}

As illustrated in Fig.~\ref{fig:pipeline}, $\pi$ reads the most recent observation $\mathcal{O}_{t_1-m_o+1:t_1}$ at time $t_1$, and after an inference duration of $t_{\text{Inf}}$, generates an action sequence $\mathbf{u}_{t_1+1:t_1+m_a}$. The action manager at time \( t_2 \approx t_1 + t_{\text{Inf}} \) must merge $\mathbf{u}_{t_1+1:t_1+m_a}$ with current action buffer $\mathbf{\mathcal{A}}$, where ``\( \approx \)'' indicates that \( t_2 \) is the first execution moment after obtaining the action sequence generated from the observation at \( t_1 \). A critical issue of merging lies in determining the correspondence between actions in $\mathbf{u}_{t_1+1:t_1+m_a}$ and those already in $\mathbf{\mathcal{A}}$. Aligning them by timestamp is a straightforward approach, treating the action at \( t_2 \) in \(\mathbf{u}_{t_1+1:t_1+m_a}\) as corresponding to the action at \( t_2 \) in $\mathbf{\mathcal{A}}$. However, in practice, we observe that due to inconsistencies in policy inference, actions predicted for the same timestamp may not match, potentially leading to action backtracking, i.e., repeatedly executing previously completed actions. To address this, we use timestamp alignment as an initial solution and select the appropriate match by comparing the similarity of actions as follows:
\begin{equation}
\begin{split} 
    t_u = \mathop{\arg\min}_{t} \frac{\sum_{k \in \mathbb{K}} d(u_{t+k}, A_{t_2+k})}{|\mathbb{K}|}, \\
    t\in[t_2-m_s, t_2+m_s], 
\end{split} 
\end{equation}
where \( d() \) denotes the Euclidean distance, \( \mathbb{K} \) represents the overlapping range between $\mathbf{u}_{t_1+1:t_1+m_a}$ and $\mathbf{\mathcal{A}}$, \( |\mathbb{K}| \) is the cardinality of \( \mathbb{K} \), and \( m_s \) is the predefined search range. The resulting \( t_u \) indicates that \( \textbf{u}_{t_u:t_u+k} \) corresponds to \( \mathcal{A}_{t_2:t_2+k} \).

After establishing the correspondence, inspired by ACT~\cite{ACT}, we employ temporal ensemble to further enhance the smoothness of the actions. Specifically, the action buffer \( \mathcal{A} \) is updated as follows:

\begin{equation}
A_{t_2+k} =
\begin{cases}
w^{\text{old}}_{t} \cdot A_{t_2+k} + w^{\text{new}}_t \cdot u_{t_u +k}, & 0 \leq k \leq |\mathcal{A}| - t_2, \\
u_{t_u +k}, & k > |\mathcal{A}| - t_2,
\end{cases}
\end{equation}

\noindent where the weights $w^{\text{new}}_{t}$ of $\mathbf{u}_{t_1+1:t_1+m_a}$ are computed following an exponential distribution parameterized by the hyper parameter $\alpha$. For the action manager, the weight of each time step is initialized to 0. After performing temporal ensemble, the weight of each time step is updated by adding the corresponding weight of the predicted action, i.e.:
\begin{align}
w^{\text{new}}_{t+i} &= \exp(-\alpha i), i \in [1, m_a], \\
w^{\text{old}}_t &= w^{\text{old}}_t + w^{\text{new}}_{t}.
\end{align}

\subsubsection{Action Interpolation}
Additionally, considering the relatively low frequency of the policy network's action sequence output compared to the high frequency required for actual action execution, we perform linear interpolation for actions at finer time step $\tau$. Specifically, for $\tau$ located between $t$ and $t+1$, the interpolated action can be expressed as:

\begin{equation}
A_{\tau} = (\tau - t) A_{t} + (t+1 - \tau) A_{t+1}.
\end{equation}
The continuous merging of action sequences into the action buffer ensures that interpolation is always performed internally, avoiding the risks associated with external extrapolation.

\subsection{Prediction Compensation Based Stabilizer}
\label{sec:stabilizer}
After obtaining high-frequency actions in real time through the action manager, we still need to address the impact of the robot's motion on manipulation, such as the robot's active movement or the wobble caused by its instability.
The most direct approach involves monitoring the robot's motion and using this data to refine its arm actions. 
However, considering computation, communication and control delays, this method adjusts the robotic arm's action using outdated robot position data, which significantly reduces the success rate of high-precision manipulation tasks.
Therefore, we design a prediction compensation based stabilizer, which predicts the platform's motion over a short future period and modifies the actions to counteract the motion effectively.

\begin{figure}[t]
	\centering
	\includegraphics[width = .9\linewidth]{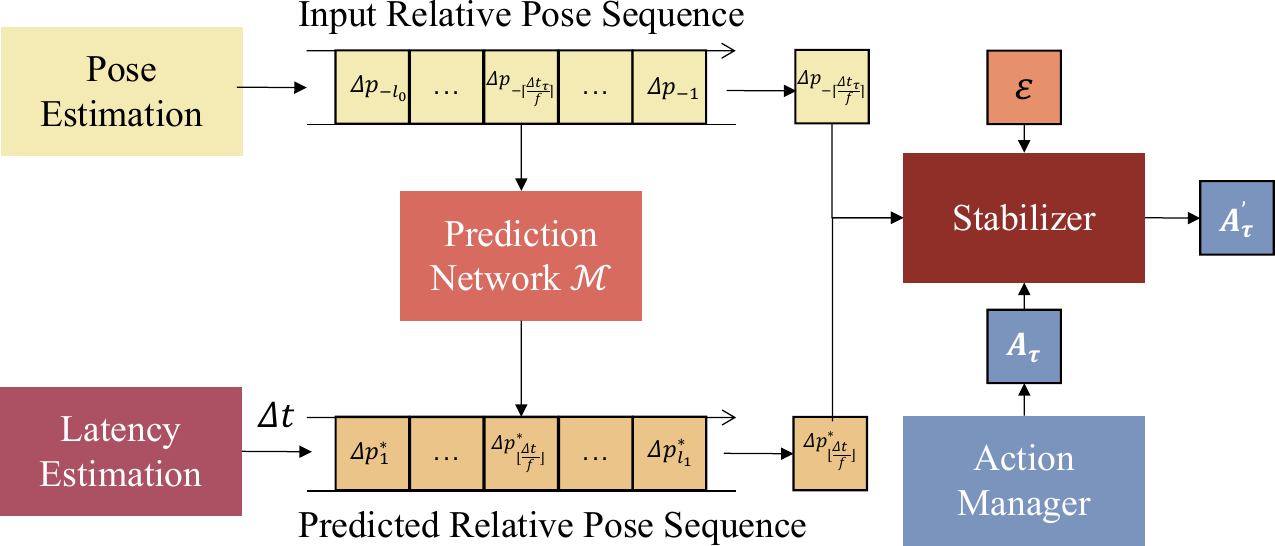}
	\caption{By utilizing the pose at action generation, the predicted pose, and the extrinsic parameters between the robotic arm and the pose estimator as inputs, the stabilizer compensates for each action output by the action manager, whether derived from the policy network or interpolation,  thereby counteracting platform motion.}
	\label{fig:stabilizer}
\end{figure}

\begin{figure*}[t]  
    \centering   
    \begin{minipage}[t]{0.24\linewidth}  
        \centering  
        \subfigure[\sysnameA] {\includegraphics[width=\linewidth]{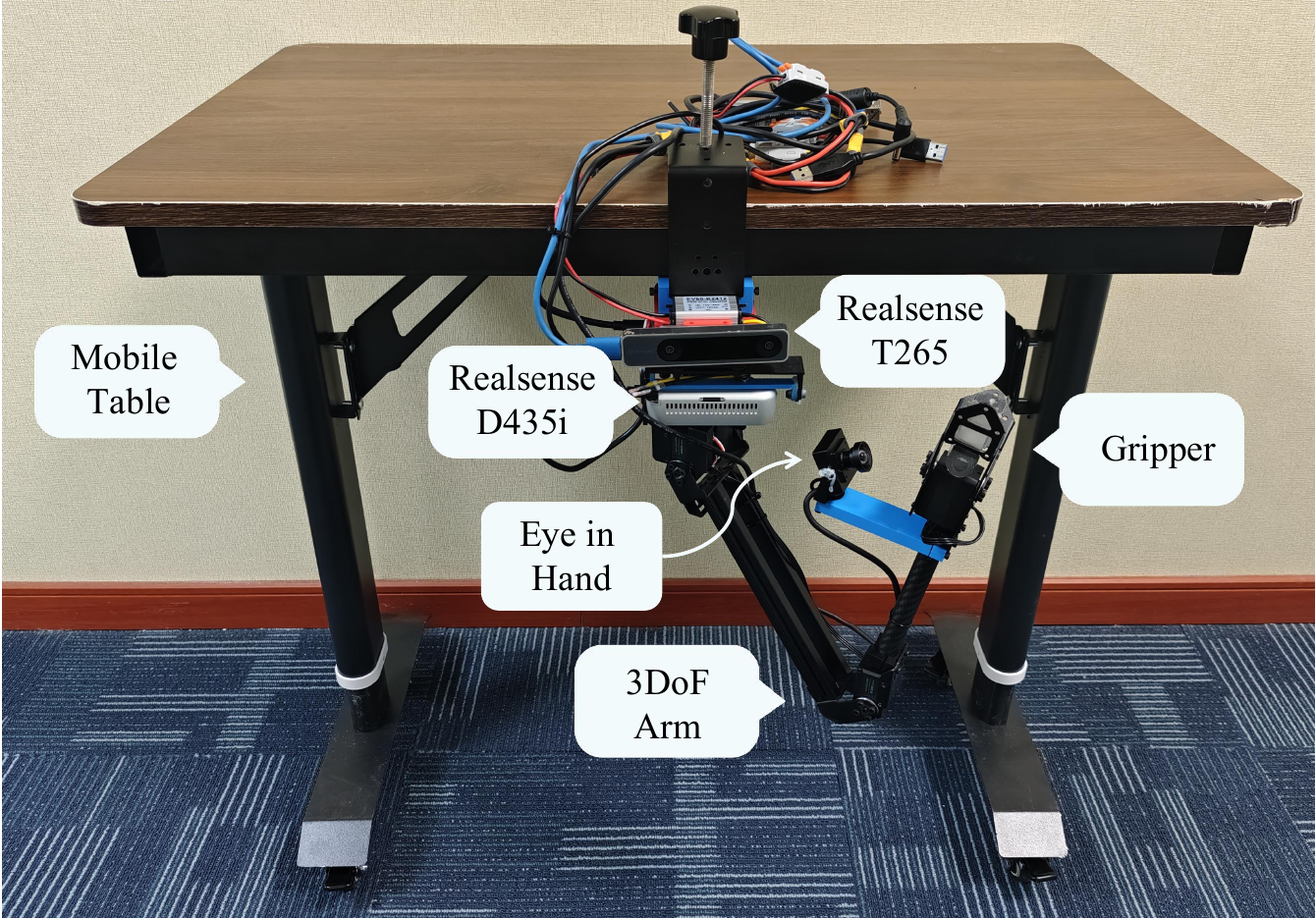}} 
    \end{minipage}  
    \hfill   
    \begin{minipage}[t]{0.24\linewidth}  
        \centering  
        \subfigure[\sysnameB] {\includegraphics[width=\linewidth]{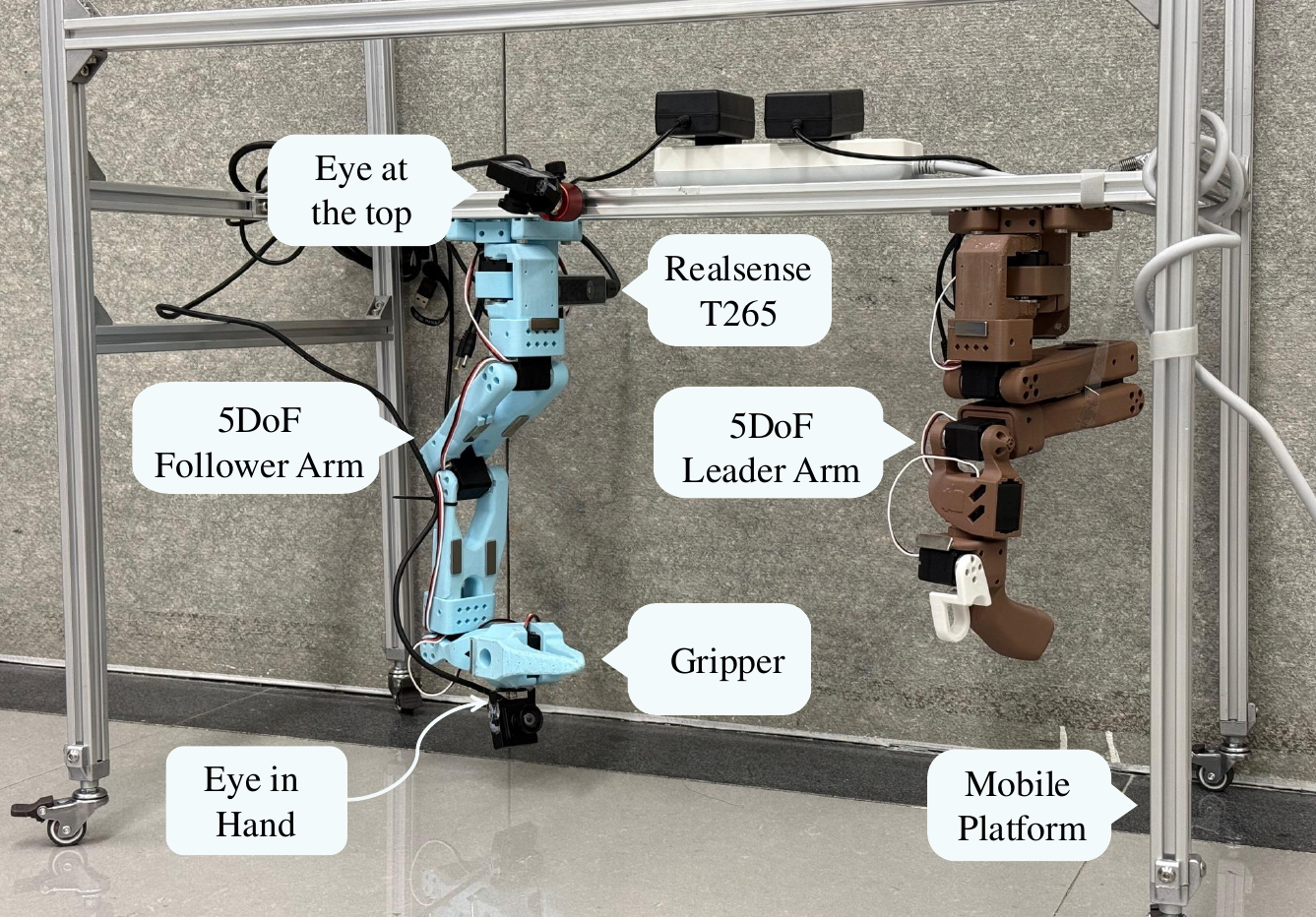}}  
    \end{minipage}  
    \hfill   
    \begin{minipage}[t]{0.24\linewidth}  
        \centering  
        \subfigure[\sysnameD] {\includegraphics[width=\linewidth]{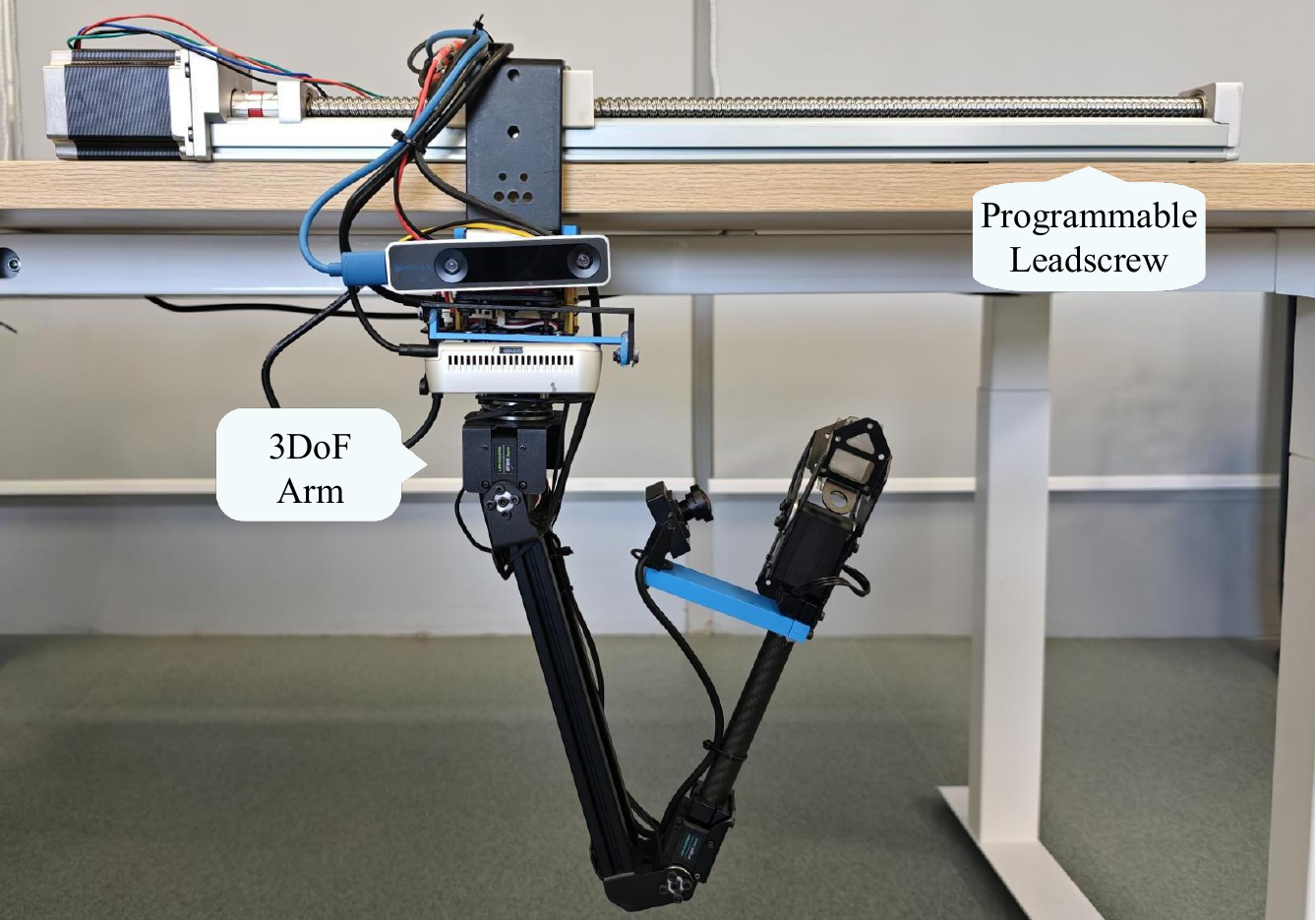}}  
    \end{minipage} 
    \hfill   
    \begin{minipage}[t]{0.24\linewidth}  
        \centering  
        \subfigure[\sysnameC] {\includegraphics[width=\linewidth]{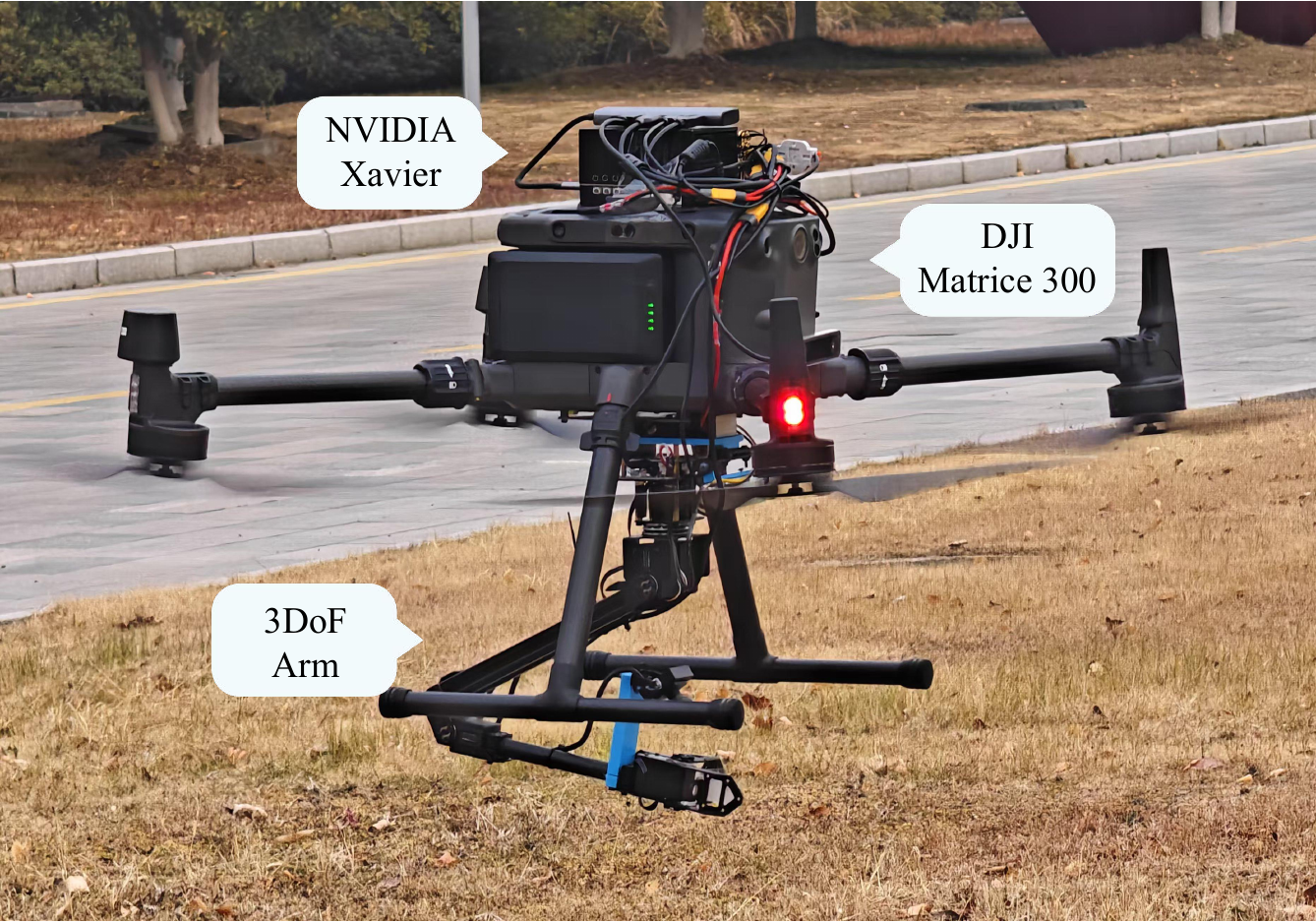}}  
    \end{minipage}
    \caption{Four experimental configurations we design.  
Each is equipped with two cameras for observation inputs and a T265 camera for pose estimation. The robotic arms operate independently of the robot platforms, with \sysnameA, \sysnameD and \sysnameC sharing the same arm. } 
    \label{fig:system_platforms}
\end{figure*}
\begin{figure*}[t]
	\centering
	\includegraphics[width = .9\linewidth]{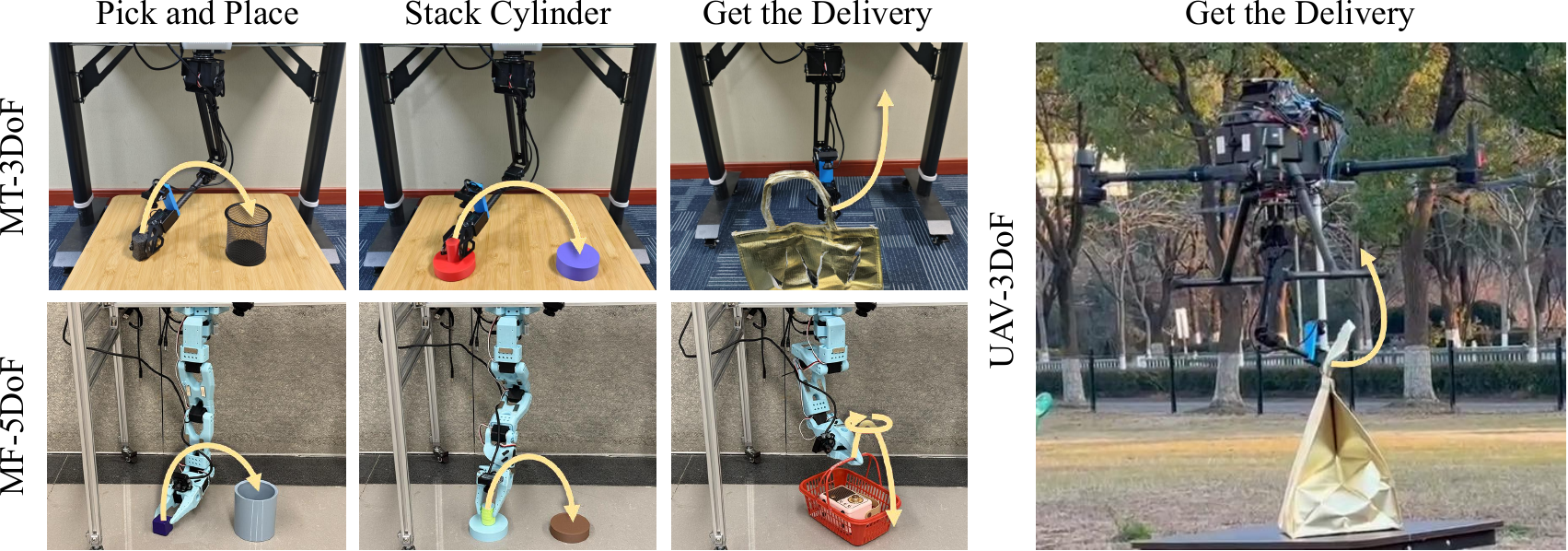}
	\caption{ 
We design three tasks of varying difficulty levels and conduct experiments across three  experimental configurations.}
	\label{fig:tasks}
\end{figure*}

As shown in Fig.~\ref{fig:stabilizer}, we first implement a pose prediction module incorporating two parts: a SLAM module for pose estimation and a lightweight prediction network. In terms of data flow, we employ a real-time, high-frequency visual SLAM using stereo camera with an IMU\footnote{https://www.intelrealsense.com/visual-inertial-tracking-case-study/} to obtain a high-frequency sequence of poses. Then, for each prediction in prediction network $\mathcal{M}$, we use the current pose $p_0$ as the base coordinate system and calculate the sequence of high-frequency relative pose $\Delta P_{-l_0:-1}=\{\Delta p_{-l_0},...,\Delta p_{-1}\}$ from the past $l_0$ frames, where $\Delta p_{-i} = p_{0}^{-1} \cdot p_{-i}$ and $\Delta P_{-l_0:-1}$ serves as the input to the prediction network. The prediction network directly outputs the future $l_1$ frames' high-frequency relative pose sequence $\Delta P^*_{1:l_1}=\{\Delta p^*_{1},...,\Delta p^*_{l_1}\}$, which predicts the future pose by $p^*_i=p_0 \cdot \Delta p^*_{i}$. 

Considering the limitation of computational resources and the requirements for system real-time performance, we implement a lightweight prediction network $\mathcal{M}$. Specifically, we parallelize an LSTM~\cite{lstm} and a GRU~\cite{gru}, which process the same input. We concatenate their corresponding temporal outputs along the feature dimension and finally pass them through a fully connected layer to produce the prediction results, that is:
\begin{equation}
\Delta P^*_{1:l_1} = \mathcal{M}(\Delta P_{-l_0:-1}),
\end{equation}
\begin{equation}
\mathcal{M}(x) = \text{FC}(\text{concat}(\text{LSTM}(x), \text{GRU}(x))).
\end{equation}

Ultimately, the action offset caused by robot motion for the target action \( A_\tau \) can be expressed as:
\begin{equation}
\delta(\tau, \Delta t, \mathcal{E}) = 
\mathcal{E}^{-1} \cdot (
\Delta p^{-1}_{\lceil-\Delta t_\tau / f \rceil} \cdot 
\Delta p^*_{\lfloor\Delta t / f \rfloor})^{-1} \cdot 
\mathcal{E},
\end{equation}
where $\mathcal{E}$ denotes the extrinsics between the visual SLAM coordinate system and the arm coordinate system, manually pre-calibrated; \( \Delta t_\tau \) represents the  time interval between the generation of \( A_\tau \) by policy network \( \pi \) and its execution; \( f \) indicates the output frequency of the SLAM system. \( \Delta t \), which will be estimated online in Sec.~\ref{sec:latency_estimation}, is the total latency spanning from pose estimation to action execution, incorporating processing time for pose estimation, pose prediction, data transmission, and action execution. The compensated action ultimately executed by the robot becomes
\begin{equation}
A'_{\tau} =\delta(\tau, \Delta t, \mathcal{E}) \cdot A_{\tau}.
\end{equation}



\subsection{Latency Estimation}
\label{sec:latency_estimation}
To align \sysname's optimization objectives with task execution goals, we implement a pre-mission warm-up routine to perform online calibration of system latency, thereby enabling the stabilizer module to accurately adjust the actions.
To achieve this, the stabilizer module is employed to help the robot do the end-effector stabilization task, enabling quantitative assessment of various latencies' impact on operational stability. The gripper maintains hold of a red  marker while a fixed monocular third-view camera tracks its positional variance during robotic arm base moving.
We conduct a linear search for the latency starting from zero, identifying the latency that minimizes the marker's movement as the system delay. To ensure that variations of the robot’s movement don't affect the correctness of the latency estimation, we adopt the velocity ratio between the marker's pixel-plane motion and the VSLAM-estimated platform motion as the evaluation metric.

\section{Implementation}
\subsection{Hardware}
\label{sec:hardware}

Our \sysname is a modular mobile manipulation system, which consists of three core components: 1) a reconfigurable mobile base platform, 2) a multi-DoF robot arm, and 3) a computing platform. For imitation learning policies inputs, the arm is equipped with two RGB cameras: one near the end effector to monitor interaction dynamics and another at the arm base to observe objects of interest. A forward-facing camera is additionally integrated to support real-time pose estimation.

To validate system versatility, as illustrated in Fig.~\ref{fig:system_platforms}, we implement four experimental configurations:
\begin{itemize}
    \item \textbf{\sysnameA}: Mobile Table with 3-DoF arm\footnote{https://www.waveshare.com/product/robotics/robot-arm-control/robot-arm/roarm-m2-s.htm}
    \item \textbf{\sysnameB}:  Mobile aluminum  frame with 5-DoF arm \footnote{https://github.com/TheRobotStudio/SO-ARM100}
    \item \textbf{\sysnameD}: Programmable Leadscrew with 3-DoF arm
    \item  \textbf{\sysnameC}: DJI Matrice 300\footnote{https://www.dji.com/support/product/matrice-300} with 3-DoF arm
\end{itemize}

All configurations utilize an NVIDIA Jetson AGX Xavier Developer Kit~\footnote{https://www.nvidia.com/en-us/autonomous-machines/embedded-systems/jetson-agx-xavier/} as the computing platform. Specially, all components of \sysnameC except the drone are powered via USB interfaces from the drone.

\begin{table*}[t]
\center
\caption{Success rate of our configurations and tasks. The best in dynamic is in bold.}
\label{table:experiment_results}

\begin{tabular}{|c|c|c|c|c|c|}
\hline
Operation Mode           & Configurations             & Method       & Pick and Place  & Stack Cylinder  & Get the Delivery \\ \hline
\multirow{4}{*}{Static}  & \multirow{2}{*}{\sysnameA} & DP           & 73.3\%          & 66.7\%          & 100\%            \\ \cline{3-6} 
                         &                            & DP+\sysname  & 73.3\%          & 80\%            & 100\%            \\ \cline{2-6} 
                         & \multirow{2}{*}{\sysnameB} & ACT          & 93.3\%          & 100\%           & 100\%            \\ \cline{3-6} 
                         &                            & ACT+\sysname & 100\%           & 100\%           & 100\%            \\ \hline 
\multirow{6}{*}{Dynamic} & \multirow{2}{*}{\sysnameA} & DP           & 0\%             & 0\%             & 20\%             \\ \cline{3-6} 
                         &                            & DP+\sysname  & \textbf{53.3\%} & \textbf{53.3\%} & \textbf{93.3\%}  \\ \cline{2-6} 
                         & \multirow{2}{*}{\sysnameB} & ACT          & 6.7\%           & 13.3\%          & 26.7\%           \\ \cline{3-6} 
                         &                            & ACT+\sysname & \textbf{53.3\%} & \textbf{66.7\%} & \textbf{86.7\%}  \\ \cline{2-6}
                         & \multirow{2}{*}{\sysnameC} & DP           & \textbackslash  & \textbackslash  & 13.3\%           \\ \cline{3-6} 
                         &                            & DP+\sysname  & \textbackslash  & \textbackslash  & \textbf{40\%}    \\ \hline
\end{tabular}
\end{table*}

\subsection{Tasks Design}
\label{sec:tasks}
As shown in the Fig.~\ref{fig:tasks}, we design three tasks with varying requirements for operational precision as follows:
\begin{itemize}
    \item \textbf{Pick and Place.} We first conduct a classic task in robot manipulation, where the robot needs to pick up a block and throw it into a small container. The block is approximately 3 cm wide, and the pen holder has a diameter of about 8 cm.
    \item \textbf{Stack Cylinder.} We attempt to stack two cylinders, each with a diameter of 8 cm. This requires the robotic arm to have a control precision within 4 cm, as any deviation beyond 4 cm results in a failed stack. To facilitate gripping, we attach a cylindrical handle with a diameter of 2.5 cm to one of the cylinders.
    \item \textbf{Get the Delivery.} Finally, we attempt to complete a real-world task—picking up takeout. The robot needs to extend its gripper near the handle of a bag or basket, grasp the bag, and lift it. We successfully perform this task using a drone as well.
\end{itemize}

These tasks are validated on \sysnameA and \sysnameB. However, on \sysnameC, only the \textbf{Get the Delivery} task is tested. This limitation is that the wind generated by the drone can blow away the target objects of the other two tasks.

\section{Experiments}

\subsection{Experimental Setup}

For each task described in Sec.\ref{sec:tasks}, we gather 50 sequences of stationary state data to train the policy network, with target objects randomly placed within the robotic arm's reachable workspace. Additionally, for each configuration outlined in Sec.\ref{sec:hardware}, we collect approximately five minutes of motion data to train the pose prediction network.

In the case of \sysnameA and \sysnameB, we simulate mobile scenarios by manually shaking the platforms. To ensure fairness, we employ a double-blind protocol where the operator shaking the platform is unaware of the specific algorithm being tested and receives no visual feedback during the process, focusing solely on generating random perturbations. This methodology not only ensures impartiality but also inherently demonstrates the robustness of our approach by leveraging the natural variability of human motion.

For \sysnameD, a programmable leadscrew moves the arm base at 12 cm/s to provide perturbations. Since the motion is precisely controllable, \sysnameD provides reproducible experimental conditions, but its regular motion makes trajectory prediction straightforward. For this reason, we only employ it for ablation studies

Regarding \sysnameC, the aerial platform's inherent instability arises from its dynamic manipulation environment, including self-induced aerodynamic disturbances and limited positional precision during hovering. We leverage these natural instability characteristics rather than introducing artificial perturbations. Notably, to accommodate the manipulation's operational range constraints and the drone's finite positional holding capacity, an operator manually maintains the relative positioning between the drone and the delivery target. This operational paradigm is adopted because that fully autonomous drone navigation exceeds the scope of this study.

As for the specific network architecture, we employ ResNet-18~\cite{resnet} as vision backbone of the policy network, with input images of shape $3\times480\times640$. The observation and prediction horizons are configured differently for each method: For DP, we set the observation horizon to $2$ and action prediction horizon to $8$ at $5$Hz, utilizing a transformer~\cite{transformer} combined with DDIM~\cite{ddim} for denoising with $20$ steps during inference. For ACT, we configure the observation horizon to $1$ and action prediction horizon to $100$ at $30$Hz. In the action manager, actions are interpolated to $50$Hz, and the temporal ensemble parameter $\alpha$ is set to $1$. The pose prediction network takes a $1$s pose sequence as input and generates $0.5$s predictions, where pose frequency is $200$Hz. This configuration ensures optimal real-time performance under constraints including inference latency, communication latency, and edge device performance variability. The code framework is adapted from the lerobot codebase~\cite{cadene2024lerobot}.

\subsection{Experiment Results}

\begin{figure*}[t]
	\centering
	\includegraphics[width = .9\linewidth]{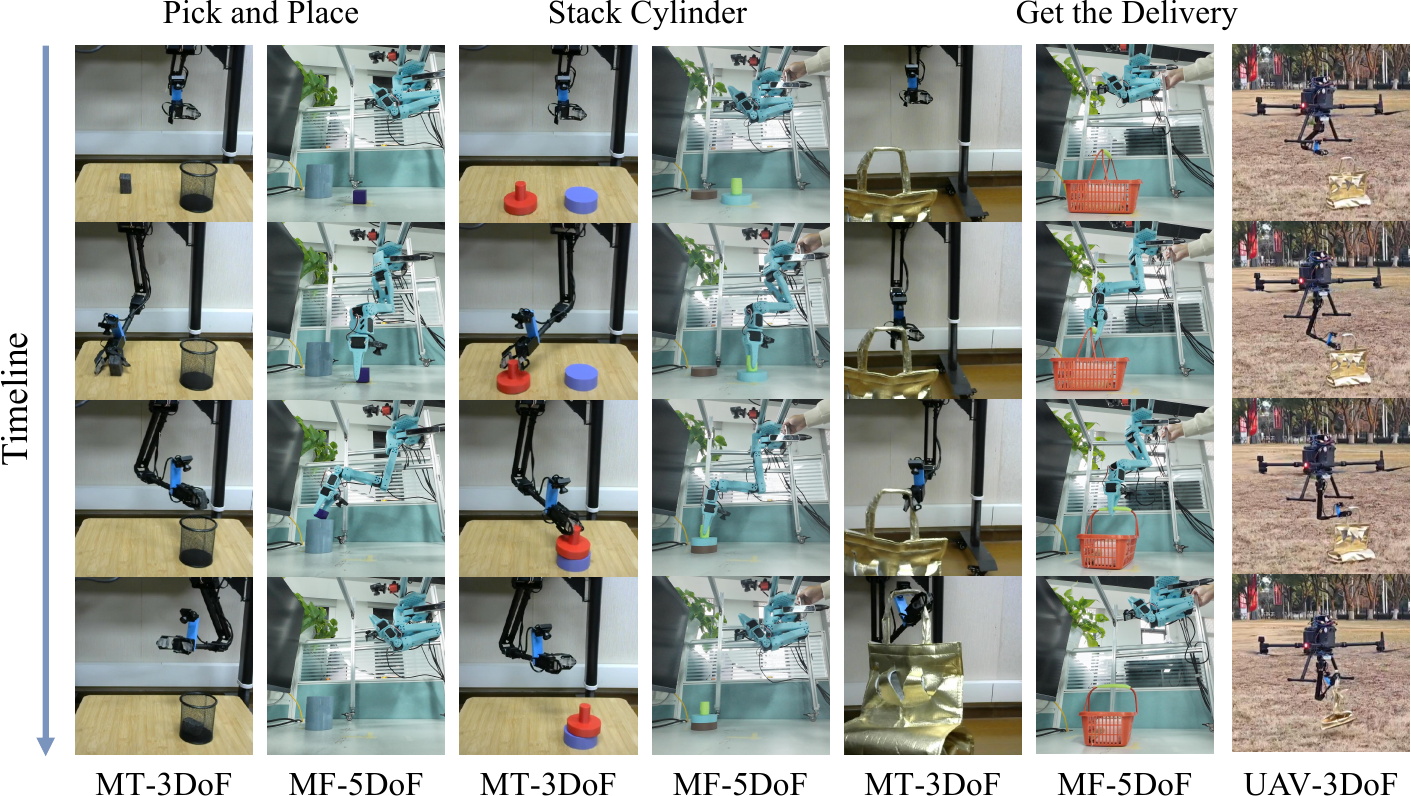}
	\caption{Temporal task progression during experiments. Beyond task completion metrics, the motion of the platforms also worth noting. Videos on the project page visually demonstrate that the \sysname enables real-time compensation for platform motion disturbances while preserving the original policy’s manipulation capabilities.}
    \vspace{-1.5em}
	\label{fig:timeline}
\end{figure*}

The experimental results presented in Tab.~\ref{table:experiment_results} demonstrate the efficacy of \sysname in enabling policy networks trained under static conditions to operate effectively on dynamically unstable platforms. Under static configurations, baseline methods (DP and ACT) achieve satisfactory performance across all tasks, \sysnameB attaining perfect scores in all tasks with ACT. However, dynamic scenarios reveal severe performance degradation for baseline methods, particularly for precision-demanding tasks like Pick and Place (up to 0\% success rate) and Stack Cylinder (up to 13.3\% success rate).

The integration of \sysname yields significant improvements, with 46.7-53.3\% relative performance recovery in dynamic Pick and Place and Stack Cylinder. Moreover, \sysnameA and \sysnameB achieved improvements of 73.3\% and 60\% in Get the delivery, respectively. Notably, the aerial platform (\sysnameC) exhibits distinct behavioral characteristics, where baseline methods achieve only 13.3\% success rate in dynamic Get the Delivery, while \sysname-enhanced policies demonstrate a 200\% performance improvement (40\% success rate). This performance gap highlights the method's adaptability to different instability sources, ranging from manual platform perturbations (\sysnameA/\sysnameB) to inherent aerodynamic uncertainties (\sysnameC).
Fig.~\ref{fig:timeline} illustrates the robotic arm's task execution process during platform movement.

It should be noted that we use two different robotic arms and two distinct policies to demonstrate STDArm's plug-and-play capability. Consequently, we did not conduct experiments to compare different policies (e.g., DP vs. ACT) on the same configurations.


\subsection{Ablation Study}

\begin{figure}[t]
	\centering
	\includegraphics[width = .9\linewidth]{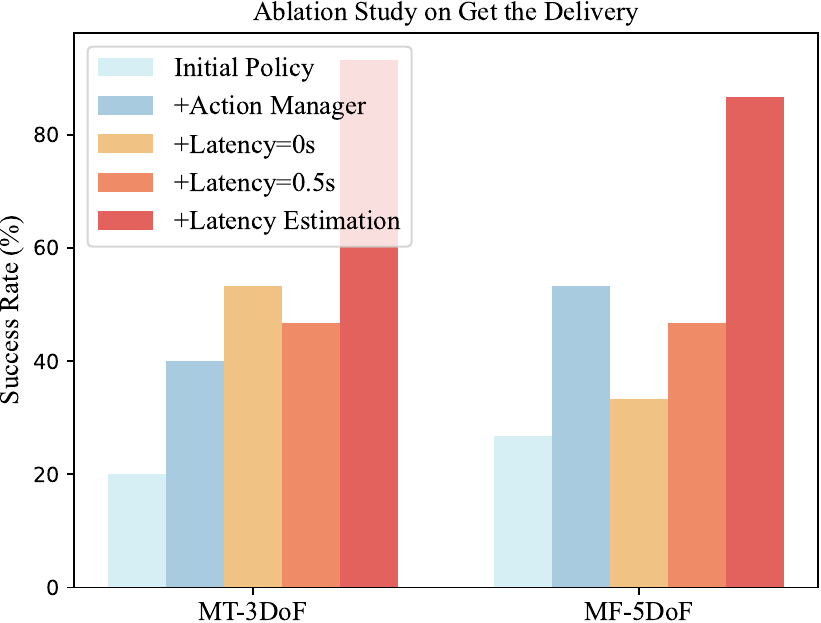}
	\caption{Ablation Study with \sysnameA and \sysnameB on Get the Delivery}
	\label{fig:ablation}
\end{figure}

To account for the potential controversy surrounding the manual influence of shaking the platform, we conduct two separate ablation studies in the Get the Delivery task. The first study employs \sysnameA and \sysnameB to evaluate the incremental contributions of the components of \sysname. The second study further systematically evaluates each component's impact across various conditions using \sysnameD.

As shown on Fig.~\ref{fig:ablation}, starting from a baseline static policy that directly employs the trained DP or ACT model without dynamic adaptation, we observe the lowest success rates (20\% on \sysnameA and 26.7\% on \sysnameB) due to its inability to handle dynamic state changes. Introducing the action manager improves performance by 20\% and 26.7\% respectively, though inference latency still limit operational stability. Subsequent integration of the stabilizer module's pose prediction mechanism with ``latency=0s'' or ``latency=0.5s'' does not noticeable improvement compare to the action manager baseline. Addressing system latency through our latency estimation module, \sysname with all components achieves peak success rates of 93.3\% on \sysnameA and 86.7\% on \sysnameB across 15 experimental trials, establishing 73.3\%/60\% absolute improvement over the initial static policy. This progressive performance gain confirms the necessity of our approach to achieve precise robotic manipulation in dynamic environments.

\setlength{\tabcolsep}{3pt}

\begin{table}[t]
\centering
\caption{Ablation study with \sysnameD on Get the Delivery.}
\label{table:ls_ablation}
\begin{tabular}{c|c|c|c|c|c}
\hline
 
\multirow{2}{*}{\makecell{No.}} & \multicolumn{2}{c|}{Action Manager}  & \multicolumn{1}{c|}{\multirow{2}{*}{Latency=0}} & \multirow{2}{*}{\begin{tabular}[c]{@{}c@{}}Latency\\ Estimation\end{tabular}} & \multirow{2}{*}{\makecell{Success\\ Rate(\%)}} \\ \cline{2-3}
& \multicolumn{1}{c|}{\makecell{Temporal\\ Ensemble}} & \multicolumn{1}{c|}{\makecell{Action\\ Interpolation}} & \multicolumn{1}{c|}{} &  &  \\

\hline

0       &                  &              &          &                   & 0            \\ 
1       & \ding{51}                &              &          &                   & 20           \\
2       &                  &              &          & \ding{51}                 &    40        \\
3       & \ding{51}                & \ding{51}            &          &                   & 26.6        \\
4       & \ding{51}                &              & \ding{51}        &                   & 33.3         \\
5       & \ding{51}                & \ding{51}            & \ding{51}        &                   & 33.3         \\
6       & \ding{51}                &              &          & \ding{51}                 & 53.3         \\
7       & \ding{51}                & \ding{51}            &          & \ding{51}                 & 80    \\

\hline

\end{tabular}
\vspace{-3mm}
\end{table}

For the second ablation study detailed in Tab.~\ref{table:ls_ablation}, each group involves 15 trials using DP as the base policy. Results underscore three key elements: temporal ensemble (No.0 vs. No.1 \& No.2 vs. No.6), latency estimation (No.5 vs. No.7), and action interpolation (No.6 vs. No.7). Ablations demonstrate that continuous decision-making and high-frequency control generated by the action manager, along with latency-aware action compensation, are essential components. 

\subsection{Accuracy of End Hold}

\begin{figure}[t]  
    \centering   
    \begin{minipage}[t]{0.48\linewidth}  
        \centering  
        \subfigure[End hold task] {\includegraphics[width=\linewidth]{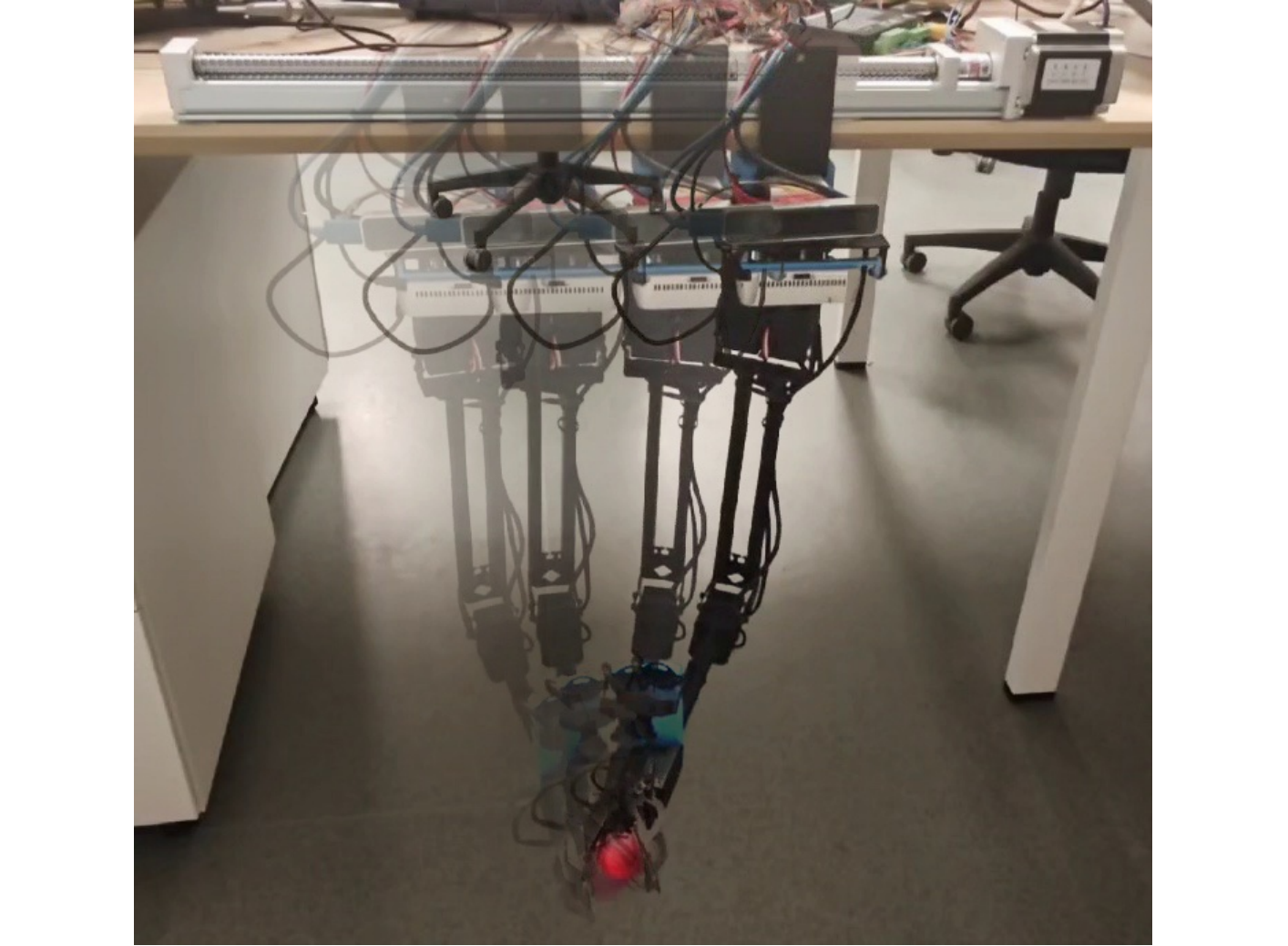}} 
    \end{minipage}  
    \hfill   
    \begin{minipage}[t]{0.48\linewidth}  
        \centering  
        \subfigure[Without correction] {\includegraphics[width=\linewidth]{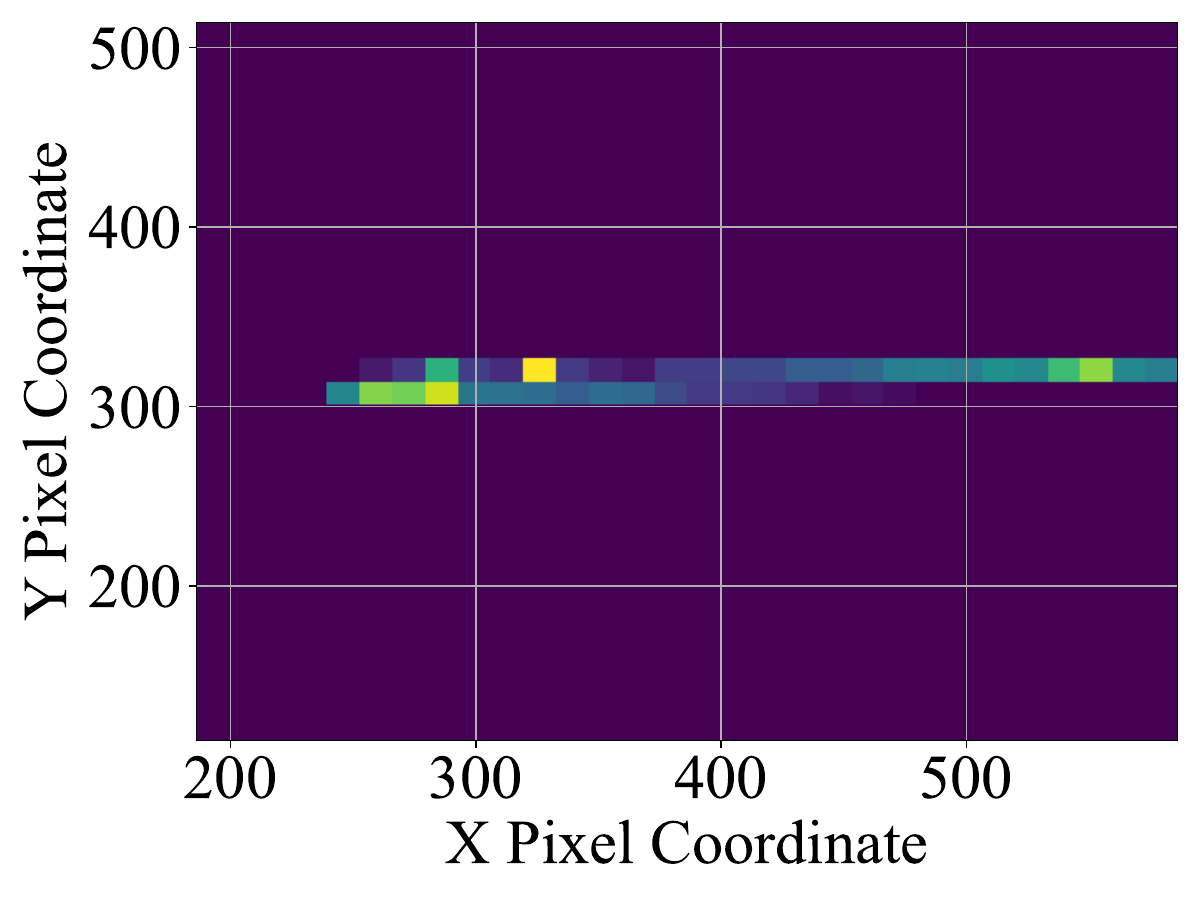}}  
    \end{minipage}  

    \begin{minipage}[t]{0.48\linewidth}  
        \centering  
        \subfigure[Latency=0s] {\includegraphics[width=\linewidth]{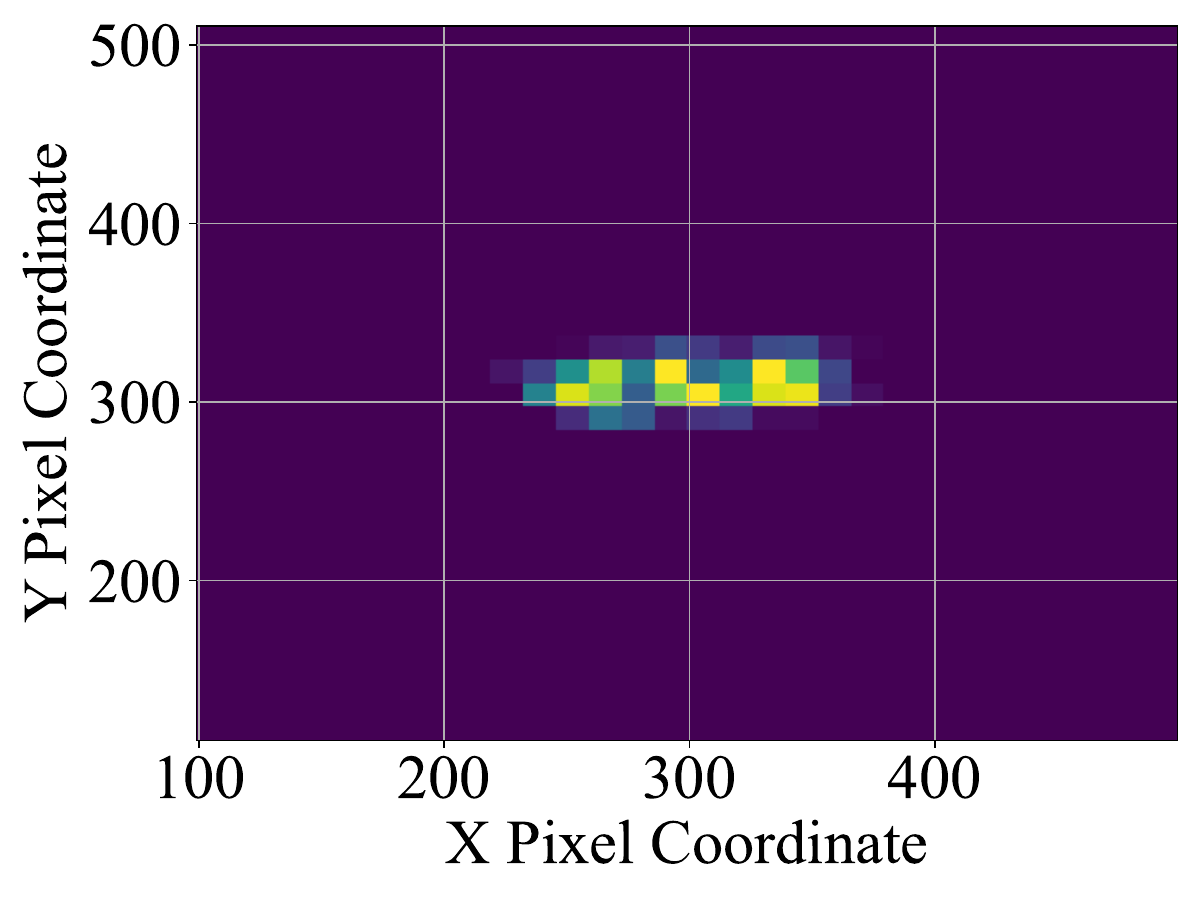}} 
    \end{minipage}  
    \hfill   
    \begin{minipage}[t]{0.48\linewidth}  
        \centering  
        \subfigure[\sysname] {\includegraphics[width=\linewidth]{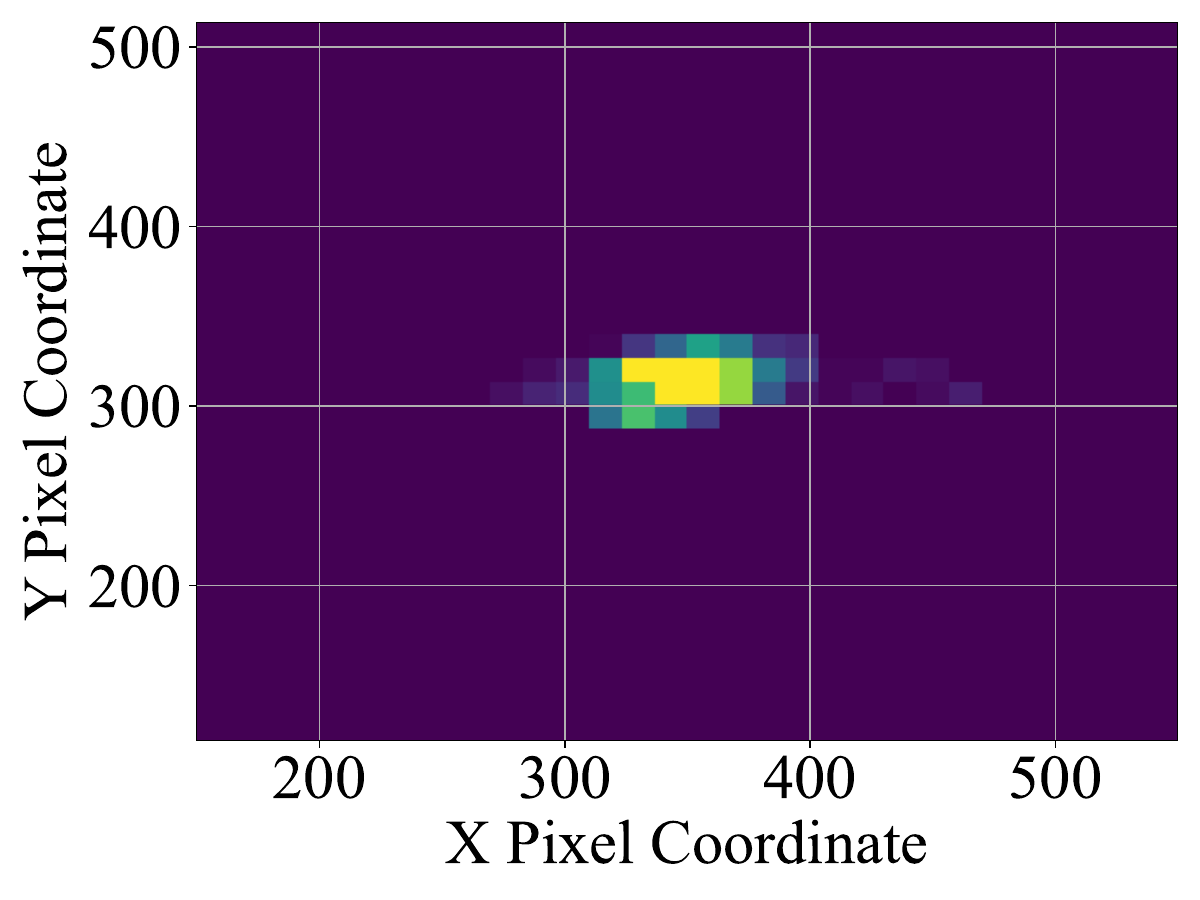}}  
    \end{minipage}

    \caption{The visualization of the end hold task and the distribution heatmaps of the red marker.}  
    \label{fig:heatmaps}
    \vspace{-1.5em}
\end{figure}

To more intuitively demonstrate the contribution of our method to end-effector stability, we conduct an end-effector holding task on \sysnameA to visualize performance. As depicted in Fig.~\ref{fig:heatmaps}(a), the robot's base is swayed left and right along with the table for 30 seconds while tracking the movement trajectory of a red ball held in the gripper. As shown in Fig.~\ref{fig:heatmaps}(b), without any corrective actions, the distribution of the red ball mirrors the movement of the manipulation, exhibiting a linear pattern. When employing the ``latency=0s'' correction, the situation improved, as seen in Fig.~\ref{fig:heatmaps}(c). However, due to the lack of prediction of the base’s movement, the corrections to the manipulation’s actions are delayed, resulting in a relatively large range of end-effector movement. Finally, with the integration of our pose prediction module, the delay from pose changes to manipulation control is effectively mitigated, as illustrated in Fig.~\ref{fig:heatmaps}(d). This advancement significantly enhanced the stability of the end-effector.


\section{Conclusion}

In this work, we introduce \sysname, a system that effectively transfers visuomotor policies trained under static data to dynamic robotic platforms. By addressing key challenges such as decision-making latency, platform-induced motion disturbances, and asynchronous control loops, \sysname significantly enhances manipulation stability during robot movement. Our design integrates an action manager for high-frequency control, a lightweight prediction network to bridge perception-action delays, and an end-effector stabilizer that ensures precise execution independent of the robot’s motion.

Through extensive evaluations across multiple robotic arms, mobile platforms, and manipulation tasks, we demonstrated that \sysname achieves centimeter-level precision, even in highly dynamic conditions. Our results confirm that \sysname enables real-time compensation for motion disturbances, achieving success rates comparable to static-trained policies under ideal conditions. This validates its effectiveness as a cost-efficient and generalizable solution for deploying visuomotor policies across diverse mobile robots.

\textbf{Limitations and future work}
A key assumption of all the contributions presented in this paper is that the foundational policy network is functional. However, during a robot's movement, inevitable changes in the camera's perspective, the background of the target object, and other factors pose significant challenges to the policy network's generalization. Our method does not address this issue. In the future, we will focus on specifically enhancing the generalization capability of the policy network to improve its performance in challenging dynamic scenarios.

\section*{Acknowledgments}
We gratefully acknowledge Prof. Chong Li from OUC and Yu Liu from USTC  for their expert advice on control theory.
This work was supported by the National Natural Science Foundation of China (No. 62332016) and the Key Research Program of Frontier Sciences, CAS (No. ZDBS-LY-JSC001).

\bibliographystyle{unsrtnat}
\bibliography{references}

\begin{thebibliography}{37}
\providecommand{\natexlab}[1]{#1}
\providecommand{\url}[1]{\texttt{#1}}
\expandafter\ifx\csname urlstyle\endcsname\relax
  \providecommand{\doi}[1]{doi: #1}\else
  \providecommand{\doi}{doi: \begingroup \urlstyle{rm}\Url}\fi

\bibitem[Urain et~al.(2024)Urain, Mandlekar, Du, Shafiullah, Xu, Fragkiadaki, Chalvatzaki, and Peters]{urain2024deep}
Julen Urain, Ajay Mandlekar, Yilun Du, Mahi Shafiullah, Danfei Xu, Katerina Fragkiadaki, Georgia Chalvatzaki, and Jan Peters.
\newblock Deep generative models in robotics: A survey on learning from multimodal demonstrations.
\newblock \emph{arXiv preprint arXiv:2408.04380}, 2024.

\bibitem[Rahmatizadeh et~al.(2018)Rahmatizadeh, Abolghasemi, B{\"o}l{\"o}ni, and Levine]{rahmatizadeh2018vision}
Rouhollah Rahmatizadeh, Pooya Abolghasemi, Ladislau B{\"o}l{\"o}ni, and Sergey Levine.
\newblock Vision-based multi-task manipulation for inexpensive robots using end-to-end learning from demonstration.
\newblock In \emph{2018 IEEE international conference on robotics and automation (ICRA)}, pages 3758--3765. IEEE, 2018.

\bibitem[Zhao et~al.(2023)Zhao, Kumar, Levine, and Finn]{ACT}
Tony~Z Zhao, Vikash Kumar, Sergey Levine, and Chelsea Finn.
\newblock Learning fine-grained bimanual manipulation with low-cost hardware.
\newblock \emph{{Robotics: Science and Systems (RSS)}}, 2023.

\bibitem[Chi et~al.(2023)Chi, Xu, Feng, Cousineau, Du, Burchfiel, Tedrake, and Song]{dp}
Cheng Chi, Zhenjia Xu, Siyuan Feng, Eric Cousineau, Yilun Du, Benjamin Burchfiel, Russ Tedrake, and Shuran Song.
\newblock Diffusion policy: Visuomotor policy learning via action diffusion.
\newblock \emph{The International Journal of Robotics Research}, page 02783649241273668, 2023.

\bibitem[Weng et~al.(2022)Weng, Bajracharya, Wang, Agrawal, and Held]{weng2022fabricflownet}
Thomas Weng, Sujay~Man Bajracharya, Yufei Wang, Khush Agrawal, and David Held.
\newblock Fabricflownet: Bimanual cloth manipulation with a flow-based policy.
\newblock In \emph{Conference on Robot Learning}, pages 192--202. PMLR, 2022.

\bibitem[Seo et~al.(2023)Seo, Han, Sim, Bang, Gonzalez, Sentis, and Zhu]{seo2023deep}
Mingyo Seo, Steve Han, Kyutae Sim, Seung~Hyeon Bang, Carlos Gonzalez, Luis Sentis, and Yuke Zhu.
\newblock Deep imitation learning for humanoid loco-manipulation through human teleoperation.
\newblock In \emph{2023 IEEE-RAS 22nd International Conference on Humanoid Robots (Humanoids)}, pages 1--8. IEEE, 2023.

\bibitem[Ha et~al.(2024)Ha, Gao, Fu, Tan, and Song]{umi_on_legs}
Huy Ha, Yihuai Gao, Zipeng Fu, Jie Tan, and Shuran Song.
\newblock Umi on legs: Making manipulation policies mobile with manipulation-centric whole-body controllers.
\newblock \emph{arXiv preprint arXiv:2407.10353}, 2024.

\bibitem[Fu et~al.(2024{\natexlab{a}})Fu, Zhao, Wu, Wetzstein, and Finn]{humanplus}
Zipeng Fu, Qingqing Zhao, Qi~Wu, Gordon Wetzstein, and Chelsea Finn.
\newblock Humanplus: Humanoid shadowing and imitation from humans.
\newblock \emph{arXiv preprint arXiv:2406.10454}, 2024{\natexlab{a}}.

\bibitem[Ze et~al.(2024)Ze, Zhang, Zhang, Hu, Wang, and Xu]{3ddp}
Yanjie Ze, Gu~Zhang, Kangning Zhang, Chenyuan Hu, Muhan Wang, and Huazhe Xu.
\newblock 3d diffusion policy: Generalizable visuomotor policy learning via simple 3d representations.
\newblock In \emph{Robotics: Science and Systems (RSS)}, 2024.

\bibitem[Kim et~al.(2024)Kim, Pertsch, Karamcheti, Xiao, Balakrishna, Nair, Rafailov, Foster, Lam, Sanketi, et~al.]{OpenVLA}
Moo~Jin Kim, Karl Pertsch, Siddharth Karamcheti, Ted Xiao, Ashwin Balakrishna, Suraj Nair, Rafael Rafailov, Ethan Foster, Grace Lam, Pannag Sanketi, et~al.
\newblock Openvla: An open-source vision-language-action model.
\newblock \emph{arXiv preprint arXiv:2406.09246}, 2024.

\bibitem[Black et~al.(2024)Black, Brown, Driess, Esmail, Equi, Finn, Fusai, Groom, Hausman, Ichter, et~al.]{black2024pi_0}
Kevin Black, Noah Brown, Danny Driess, Adnan Esmail, Michael Equi, Chelsea Finn, Niccolo Fusai, Lachy Groom, Karol Hausman, Brian Ichter, et~al.
\newblock $\pi_0$: A vision-language-action flow model for general robot control.
\newblock \emph{arXiv preprint arXiv:2410.24164}, 2024.

\bibitem[Liu et~al.(2024{\natexlab{a}})Liu, Wu, Li, Tan, Chen, Wang, Xu, Su, and Zhu]{RDT}
Songming Liu, Lingxuan Wu, Bangguo Li, Hengkai Tan, Huayu Chen, Zhengyi Wang, Ke~Xu, Hang Su, and Jun Zhu.
\newblock Rdt-1b: a diffusion foundation model for bimanual manipulation.
\newblock \emph{arXiv preprint arXiv:2410.07864}, 2024{\natexlab{a}}.

\bibitem[Gong et~al.(2024)Gong, Ding, Lyu, Huang, Sun, Zhao, Fan, and Wang]{CARP}
Zhefei Gong, Pengxiang Ding, Shangke Lyu, Siteng Huang, Mingyang Sun, Wei Zhao, Zhaoxin Fan, and Donglin Wang.
\newblock Carp: Visuomotor policy learning via coarse-to-fine autoregressive prediction.
\newblock \emph{arXiv preprint arXiv:2412.06782}, 2024.

\bibitem[Pertsch et~al.(2025)Pertsch, Stachowicz, Ichter, Driess, Nair, Vuong, Mees, Finn, and Levine]{FAST}
Karl Pertsch, Kyle Stachowicz, Brian Ichter, Danny Driess, Suraj Nair, Quan Vuong, Oier Mees, Chelsea Finn, and Sergey Levine.
\newblock Fast: Efficient action tokenization for vision-language-action models.
\newblock \emph{arXiv preprint arXiv:2501.09747}, 2025.

\bibitem[Yu et~al.(2024)Yu, Peng, Yang, Zhang, Duan, Ji, and Zhang]{LDP}
Wenhao Yu, Jie Peng, Huanyu Yang, Junrui Zhang, Yifan Duan, Jianmin Ji, and Yanyong Zhang.
\newblock Ldp: A local diffusion planner for efficient robot navigation and collision avoidance.
\newblock In \emph{2024 IEEE/RSJ International Conference on Intelligent Robots and Systems (IROS)}, pages 5466--5472. IEEE, 2024.

\bibitem[Sridhar et~al.(2024)Sridhar, Shah, Glossop, and Levine]{nomad}
Ajay Sridhar, Dhruv Shah, Catherine Glossop, and Sergey Levine.
\newblock Nomad: Goal masked diffusion policies for navigation and exploration.
\newblock In \emph{2024 IEEE International Conference on Robotics and Automation (ICRA)}, pages 63--70. IEEE, 2024.

\bibitem[Fu et~al.(2024{\natexlab{b}})Fu, Zhao, and Finn]{mobile_aloha}
Zipeng Fu, Tony~Z Zhao, and Chelsea Finn.
\newblock Mobile aloha: Learning bimanual mobile manipulation with low-cost whole-body teleoperation.
\newblock \emph{arXiv preprint arXiv:2401.02117}, 2024{\natexlab{b}}.

\bibitem[Prasad et~al.(2024)Prasad, Lin, Wu, Zhou, and Bohg]{cp}
Aaditya Prasad, Kevin Lin, Jimmy Wu, Linqi Zhou, and Jeannette Bohg.
\newblock Consistency policy: Accelerated visuomotor policies via consistency distillation.
\newblock \emph{{Robotics: Science and Systems (RSS)}}, 2024.

\bibitem[Lu et~al.(2024)Lu, Gao, Chen, Dai, Wang, and Tang]{manicm}
Guanxing Lu, Zifeng Gao, Tianxing Chen, Wenxun Dai, Ziwei Wang, and Yansong Tang.
\newblock Manicm: Real-time 3d diffusion policy via consistency model for robotic manipulation.
\newblock \emph{arXiv preprint arXiv:2406.01586}, 2024.

\bibitem[Hu et~al.(2024)Hu, Liu, Liu, and Liu]{hu2024adaflow}
Xixi Hu, Bo~Liu, Xingchao Liu, and Qiang Liu.
\newblock Adaflow: Imitation learning with variance-adaptive flow-based policies.
\newblock \emph{arXiv preprint arXiv:2402.04292}, 2024.

\bibitem[Liu et~al.(2024{\natexlab{b}})Liu, Hamid, Xie, Lee, Du, and Finn]{bid}
Yuejiang Liu, Jubayer~Ibn Hamid, Annie Xie, Yoonho Lee, Maximilian Du, and Chelsea Finn.
\newblock Bidirectional decoding: Improving action chunking via closed-loop resampling.
\newblock \emph{arXiv preprint arXiv:2408.17355}, 2024{\natexlab{b}}.

\bibitem[Bajracharya et~al.(2020)Bajracharya, Borders, Helmick, Kollar, Laskey, Leichty, Ma, Nagarajan, Ochiai, Petersen, et~al.]{bajracharya2020mobile}
Max Bajracharya, James Borders, Dan Helmick, Thomas Kollar, Michael Laskey, John Leichty, Jeremy Ma, Umashankar Nagarajan, Akiyoshi Ochiai, Josh Petersen, et~al.
\newblock A mobile manipulation system for one-shot teaching of complex tasks in homes.
\newblock In \emph{2020 IEEE International Conference on Robotics and Automation (ICRA)}, pages 11039--11045. IEEE, 2020.

\bibitem[Chestnutt et~al.(2005)Chestnutt, Lau, Cheung, Kuffner, Hodgins, and Kanade]{chestnutt2005footstep}
Joel Chestnutt, Manfred Lau, German Cheung, James Kuffner, Jessica Hodgins, and Takeo Kanade.
\newblock Footstep planning for the honda asimo humanoid.
\newblock In \emph{Proceedings of the 2005 IEEE international conference on robotics and automation}, pages 629--634. IEEE, 2005.

\bibitem[Feng et~al.(2014)Feng, Whitman, Xinjilefu, and Atkeson]{feng2014optimization}
Siyuan Feng, Eric Whitman, X~Xinjilefu, and Christopher~G Atkeson.
\newblock Optimization based full body control for the atlas robot.
\newblock In \emph{2014 IEEE-RAS International Conference on Humanoid Robots}, pages 120--127. IEEE, 2014.

\bibitem[Khatib et~al.(1996)Khatib, Yokoi, Chang, Ruspini, Holmberg, Casal, and Baader]{khatib1996force}
Oussama Khatib, K~Yokoi, K~Chang, D~Ruspini, R~Holmberg, A~Casal, and A~Baader.
\newblock Force strategies for cooperative tasks in multiple mobile manipulation systems.
\newblock In \emph{Robotics Research: The Seventh International Symposium}, pages 333--342. Springer, 1996.

\bibitem[Wyrobek et~al.(2008)Wyrobek, Berger, Van~der Loos, and Salisbury]{wyrobek2008towards}
Keenan~A Wyrobek, Eric~H Berger, HF~Machiel Van~der Loos, and J~Kenneth Salisbury.
\newblock Towards a personal robotics development platform: Rationale and design of an intrinsically safe personal robot.
\newblock In \emph{2008 IEEE International Conference on Robotics and Automation}, pages 2165--2170. IEEE, 2008.

\bibitem[Rudin et~al.(2022)Rudin, Hoeller, Reist, and Hutter]{rudin2022learning}
Nikita Rudin, David Hoeller, Philipp Reist, and Marco Hutter.
\newblock Learning to walk in minutes using massively parallel deep reinforcement learning.
\newblock In \emph{Conference on Robot Learning}, pages 91--100. PMLR, 2022.

\bibitem[Fu et~al.(2023)Fu, Cheng, and Pathak]{fu2023deep}
Zipeng Fu, Xuxin Cheng, and Deepak Pathak.
\newblock Deep whole-body control: learning a unified policy for manipulation and locomotion.
\newblock In \emph{Conference on Robot Learning}, pages 138--149. PMLR, 2023.

\bibitem[Chen et~al.(2022)Chen, Wu, Zhang, Miao, Zhong, Zhang, and Wang]{chen2022image}
Yanjie Chen, Yangning Wu, Zhenguo Zhang, Zhiqiang Miao, Hang Zhong, Hui Zhang, and Yaonan Wang.
\newblock Image-based visual servoing of unmanned aerial manipulators for tracking and grasping a moving target.
\newblock \emph{IEEE Transactions on Industrial Informatics}, 19\penalty0 (8):\penalty0 8889--8899, 2022.

\bibitem[Cao et~al.(2023)Cao, Li, Liu, and Zhao]{cao2023eso}
Huazi Cao, Yongqi Li, Cunjia Liu, and Shiyu Zhao.
\newblock Eso-based robust and high-precision tracking control for aerial manipulation.
\newblock \emph{IEEE Transactions on Automation Science and Engineering}, 21\penalty0 (2):\penalty0 2139--2155, 2023.

\bibitem[Wang et~al.(2023)Wang, Chen, Guo, Yu, Zhang, Guo, and Wang]{wang2023millimeter}
Meng Wang, Zeshuai Chen, Kexin Guo, Xiang Yu, Youmin Zhang, Lei Guo, and Wei Wang.
\newblock Millimeter-level pick and peg-in-hole task achieved by aerial manipulator.
\newblock \emph{IEEE Transactions on Robotics}, 2023.

\bibitem[Hochreiter(1997)]{lstm}
S~Hochreiter.
\newblock Long short-term memory.
\newblock \emph{Neural Computation MIT-Press}, 1997.

\bibitem[Chung et~al.(2014)Chung, Gulcehre, Cho, and Bengio]{gru}
Junyoung Chung, Caglar Gulcehre, KyungHyun Cho, and Yoshua Bengio.
\newblock Empirical evaluation of gated recurrent neural networks on sequence modeling.
\newblock \emph{arXiv preprint arXiv:1412.3555}, 2014.

\bibitem[He et~al.(2016)He, Zhang, Ren, and Sun]{resnet}
Kaiming He, Xiangyu Zhang, Shaoqing Ren, and Jian Sun.
\newblock Deep residual learning for image recognition.
\newblock In \emph{Proceedings of the IEEE conference on computer vision and pattern recognition}, pages 770--778, 2016.

\bibitem[Vaswani(2017)]{transformer}
A~Vaswani.
\newblock Attention is all you need.
\newblock \emph{Advances in Neural Information Processing Systems}, 2017.

\bibitem[Song et~al.(2020)Song, Meng, and Ermon]{ddim}
Jiaming Song, Chenlin Meng, and Stefano Ermon.
\newblock Denoising diffusion implicit models.
\newblock \emph{arXiv preprint arXiv:2010.02502}, 2020.

\bibitem[Cadene et~al.(2024)Cadene, Alibert, Soare, Gallouedec, Zouitine, and Wolf]{cadene2024lerobot}
Remi Cadene, Simon Alibert, Alexander Soare, Quentin Gallouedec, Adil Zouitine, and Thomas Wolf.
\newblock Lerobot: State-of-the-art machine learning for real-world robotics in pytorch.
\newblock \url{https://github.com/huggingface/lerobot}, 2024.

\end{thebibliography}

\end{document}